\documentclass{article}
\usepackage{iclr2026_conference,times}
\usepackage{graphicx}
\usepackage{subcaption}
\usepackage{amsmath}
\usepackage{amssymb}
\usepackage{booktabs}
\usepackage{hyperref}
\usepackage{xcolor}

\title{Event Embedding of Protein Networks : \\Compositional Learning of Biological Function}

\iclrfinalcopy

\author{Antonin Sulc\\Lawrence Berkeley National Lab,\\{asulc@lbl.gov}}

\begin{document}

\maketitle

\begin{abstract} 
In this work, we study whether enforcing strict compositional structure in sequence embeddings yields meaningful geometric organization when applied to protein-protein interaction networks.  Using Event2Vec, an additive sequence embedding model, we train 64-dimensional representations on random walks from the human STRING interactome, and compare against a DeepWalk baseline based on Word2Vec, trained on the same walks. We find that compositional structure substantially improves pathway coherence (30.2$\times$ vs 2.9$\times$ above random), functional analogy accuracy (mean similarity 0.966 vs 0.650), and hierarchical pathway organization, while geometric properties such as norm--degree anticorrelation are shared with or exceeded by the non-compositional baseline. These results indicate that enforced compositionality specifically benefits relational and compositional reasoning tasks in biological networks.
\end{abstract}

\section{Introduction}

Protein function emerges from network context. Traditional graph embeddings (DeepWalk~\citep{deepwalk}, Node2Vec~\citep{node2vec}) learn protein representations from interaction networks, but lack compositional structure, the property that vector arithmetic reflects biological relationships. 
On the other hand, Event2Vec~\citep{sulc2025event2vec}, designed for sequential data with enforced additivity (i.e., $h_t = h_{t-1}+e_{s_t}$ and hence $h_t-e_{s_t}\approx h_{t-1}$), offers a solution: if protein interactions can be represented as meaningful event sequences, then the embedding of an interaction history should be (approximately) additive,
$h_t \approx \sum_{i=1}^{t} e_{s_i}$,
which in turn implies analogy-style linear relations of the form
$e_A - e_B + e_C \approx e_D$.

We use Event2Vec on the human interactome, asking: (1) Do biological pathways cluster more coherently than under a non-compositional baseline? (2) Does protein arithmetic transfer functional relationships? (3) Which geometric properties are specific to compositional structure, and which emerge from co-occurrence statistics alone? Our results show that enforced compositionality provides large gains for relational reasoning, while some geometric properties attributed to it in prior work are shared with standard skip-gram models.

\section{Methods}

\subsection{Data}
STRING v12.0~\citep{string} human interactome consists of 16,201 proteins, 89,234 high-confidence edges (score$\geq$700). 
To generate necessary sequences for Event2vec, we generate 10 random walks per protein (length 15), yielding $\sim$160K sequences. 

\subsection{Event2Vec Training}
To train \textsc{Event2Vec}, we represent each event type $s$ with a 64-dimensional embedding $e_s\in\mathbb{R}^{64}$ and maintain a history state $h_t$ updated by a simple additive recurrence (Euclidean variant),
$h_t = h_{t-1} + e_{s_t}$ (with optional norm clipping for stability).
Training minimizes a composite objective that (i) predicts future events from the current history state within a look-ahead window of size 5 and (ii) enforces reversibility of the update via a reconstruction penalty:
\[
\mathcal{L}
= -\sum_{\Delta=1}^{5}\log P(s_{t+\Delta}\mid h_t)
\;+\;
\lambda \left\|\bigl(h_t - e_{s_t}\bigr) - h_{t-1}\right\|_2^2 .
\]
The reconstruction term encourages the enforced additivity $h_t - e_{s_t} \approx h_{t-1}$, so that the representation of an event history composes (approximately) as a sum of its constituent event embeddings.

It is important to note that we also consider a hyperbolic variant of \textsc{Event2Vec}, where composition is performed via M\"obius (gyrovector) addition in the Poincar\'e ball. Unlike Euclidean vector addition, M\"obius addition is not strictly commutative (it is only gyrocommutative and non-associative). In the Euclidean variant, by contrast, standard addition is commutative; nevertheless, the model is still trained on ordered sequences, so directionality is preserved through the step-wise recurrent update and next-event prediction objective rather than through the algebraic non-commutativity of the space.

\subsection{DeepWalk Baseline}
To isolate the effect of compositional structure, we train a DeepWalk baseline using gensim Word2Vec (skip-gram) on the \textbf{identical} random walks with matched hyperparameters: 64 dimensions, window size 5, 100 epochs. This baseline shares Event2Vec's skip-gram objective but lacks the additive reconstruction constraint, providing a direct ablation of the compositional component.

\subsection{Evaluation Framework}
We perform six complementary analyses on both models: 

\begin{itemize}\setlength\itemsep{0pt}\setlength\parskip{0pt}\setlength\parsep{0pt}\setlength\topsep{0pt}\setlength\partopsep{0pt}
  \item \textbf{Pathway coherence}: do 10 canonical pathways cluster (PI3K--AKT, MAPK/ERK, p53, Wnt, NF-$\kappa$B, Cell Cycle, Apoptosis, DNA Repair, Ribosome, OxPhos)? (Section~\ref{sec:results:pathways})
  \item \textbf{Protein arithmetic}: does $\mathrm{EGFR}:\mathrm{AKT1}::\mathrm{INSR}:?$ retrieve RAS-family proteins?
  \item \textbf{Hub architecture}: are hubs (TP53, MYC, AKT1, EGFR, SRC, JUN, BRCA1, TNF) central?
  \item \textbf{Drug targets}: do 14 FDA-approved targets show distinctive geometric signatures?
  \item \textbf{Pathway organization}: does hierarchical clustering recover biological themes?
  \item \textbf{Embedding drift}: do cumulative compositions recover signaling trajectories?
\end{itemize}

\section{Results}

\subsection{Biological Pathways Cluster}
\label{sec:results:pathways}
Pathway coherence measures how similar proteins within the same biological pathway are to each other in the embedding space, high coherence means pathway members cluster tightly together.

\begin{figure}[t]
\centering
\includegraphics[width=\linewidth]{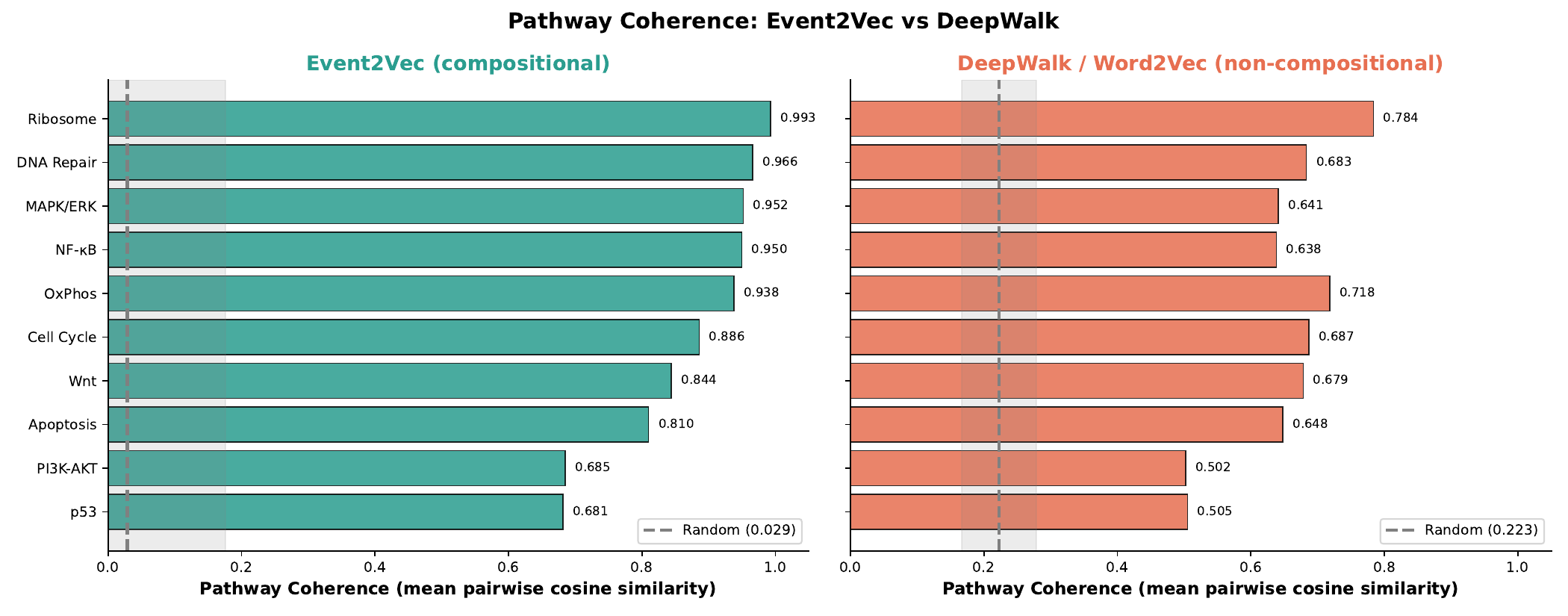}
\caption{\textbf{Pathway coherence comparison.} Event2Vec (left) shows higher coherence and much lower random baseline than DeepWalk (right) across all 10 pathways.}
\label{fig:pathways}
\end{figure}

In our experiments, known pathways show substantially higher internal coherence under Event2Vec than DeepWalk, see Figure~\ref{fig:pathways}. 
Event2Vec achieves mean pathway coherence of 0.870, or 30.2$\times$ above its random baseline of 0.029, while DeepWalk achieves 0.648, or 2.9$\times$ above its random baseline of 0.223. 
All 10 pathways show higher coherence under Event2Vec, with improvements ranging from +0.162 (Apoptosis) to +0.312 (NF-$\kappa$B). Physical complexes (Ribosome: 0.993, OxPhos: 0.938) cluster most tightly.

The large difference in random baselines is itself informative: Event2Vec's lower background similarity (0.029 vs 0.223) indicates sharper functional discrimination, with proteins from unrelated pathways placed far apart rather than in a diffuse cloud.

\subsection{Protein Arithmetic Transfers Functional Relationships}
Protein arithmetic applies vector operations to embeddings, such as computing EGFR minus AKT1 plus INSR, to test whether functional relationships learned for one protein pair transfer meaningfully to another.

\begin{figure}[t]
\centering
\includegraphics[width=\linewidth]{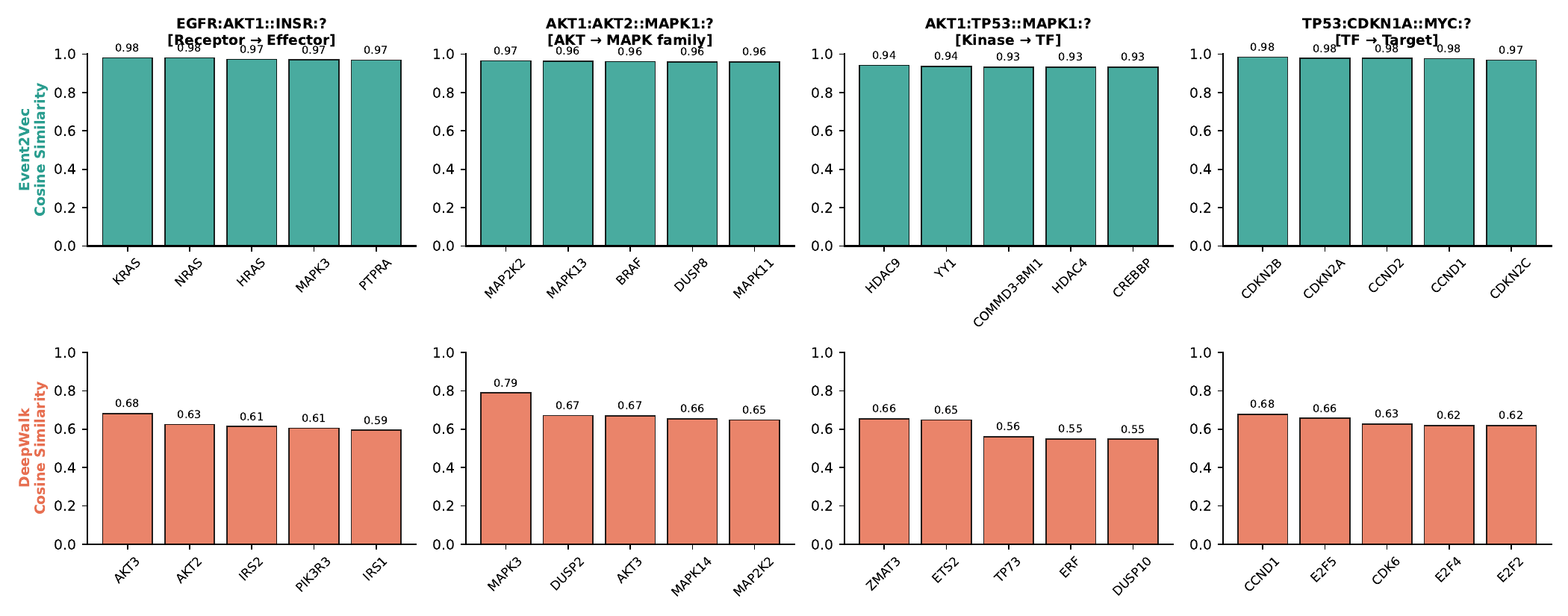}
\caption{\textbf{Protein arithmetic comparison.} Event2Vec (left column) recovers biologically specific targets with high similarity. DeepWalk (right column) returns less specific predictions with lower similarity.}
\label{fig:arithmetic}
\end{figure}

Vector arithmetic reveals the largest gap between methods (Figure~\ref{fig:arithmetic}). Event2Vec recovers biologically meaningful relationships. For instance, EGFR:AKT1::INSR:? $\rightarrow$ KRAS (0.981), correctly identifying RAS proteins in insulin signaling. DeepWalk predicts AKT3 (0.672), merely returning a protein similar to the query term rather than completing the functional analogy.

This pattern holds systematically: across 15 transfer tests spanning three functional categories (Receptor$\rightarrow$Effector, TF$\rightarrow$Target, Kinase$\rightarrow$Substrate), Event2Vec achieves mean top-1 similarity of 0.966 compared to DeepWalk's 0.650. Event2Vec scores consistently high across all tests (range: 0.93--0.99), while DeepWalk's predictions cluster around 0.65 and are often repetitive (e.g., returning AKT3 for 4/5 Receptor$\rightarrow$Effector queries). This suggests DeepWalk retrieves proteins similar to the query terms rather than capturing the directional transformation that the analogy requires.

\subsection{Network Hierarchy: Shared and Distinct Properties}

\begin{figure}[t]
\centering
\includegraphics[width=\linewidth]{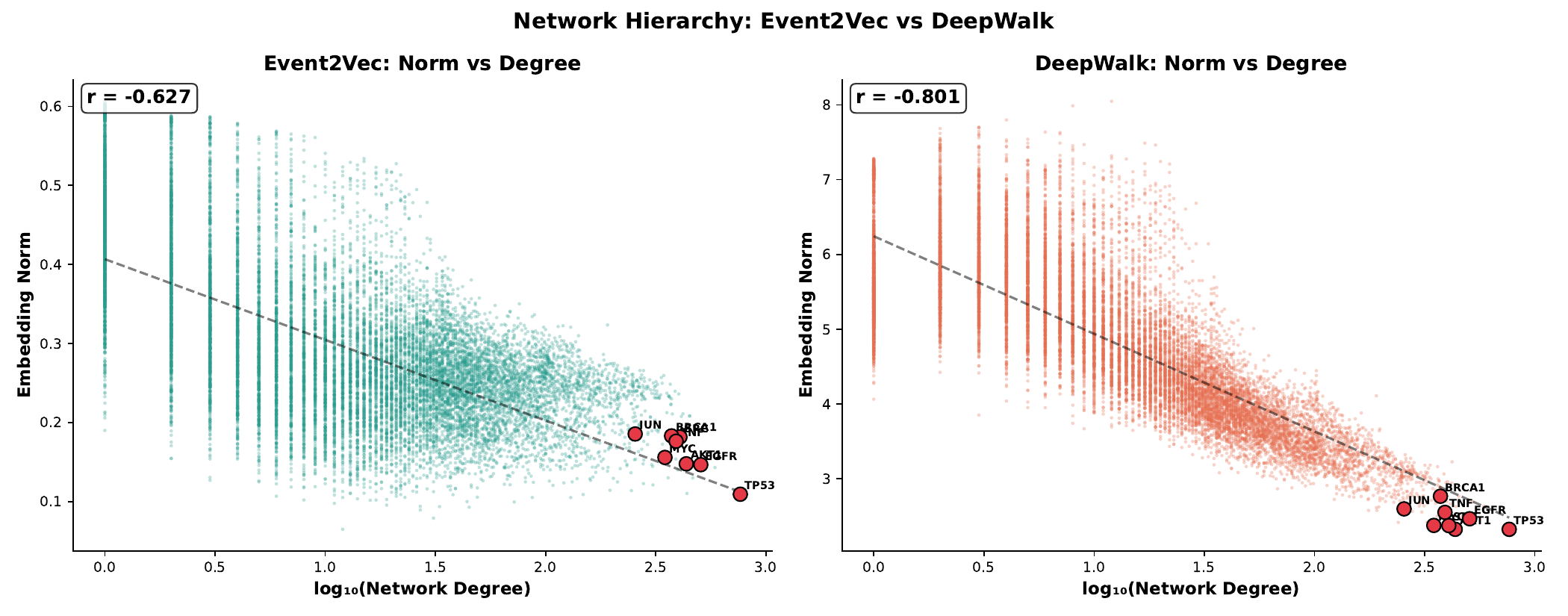}
\caption{\textbf{Norm--degree anticorrelation.} Both methods show that high-degree proteins (hubs) receive smaller embedding norms, but DeepWalk exhibits stronger anticorrelation ($r=-0.801$) than Event2Vec ($r=-0.627$). This suggests the pattern arises from skip-gram frequency effects rather than compositional structure. Red points highlight canonical hubs (TP53, MYC, AKT1, EGFR, SRC, JUN, BRCA1, TNF). Note the different norm scales: Event2Vec's additive constraint produces smaller, more tightly distributed norms.}
\label{fig:hubs}
\end{figure}

Both methods capture network hierarchy, but the comparison reveals which geometric properties arise from compositionality versus skip-gram co-occurrence statistics (Figure~\ref{fig:hubs}).

\textbf{Norm--degree anticorrelation.} High-degree proteins receive smaller embedding norms under both methods, but DeepWalk shows \emph{stronger} anticorrelation ($r=-0.801$) than Event2Vec ($r=-0.627$). This indicates that norm--degree structure is primarily a skip-gram frequency effect: proteins appearing more often in random walks are updated more frequently during training, shrinking their norms. The weaker correlation in Event2Vec may reflect the additive constraint redistributing norm across the embedding space.

\textbf{Hub centrality.} Despite the weaker norm correlation, Event2Vec places hubs 3.6$\times$ closer to the embedding centroid than random proteins ($p=1.04\times10^{-9}$), compared to 2.7$\times$ for DeepWalk ($p=1.56\times10^{-11}$). The compositional constraint thus sharpens hub--periphery organization even as it attenuates the raw norm--degree relationship.

\textbf{Drug targets.} Both methods distinguish FDA-approved drug targets from non-targets by embedding norm (Event2Vec: $p=8.41\times10^{-7}$; DeepWalk: $p=3.75\times10^{-8}$). The comparable significance suggests this separation is driven by network centrality---drug targets tend to be high-degree hubs---rather than compositional structure.

\subsection{Hierarchical Pathway Organization}

\begin{figure}[t]
\centering
\includegraphics[width=\linewidth]{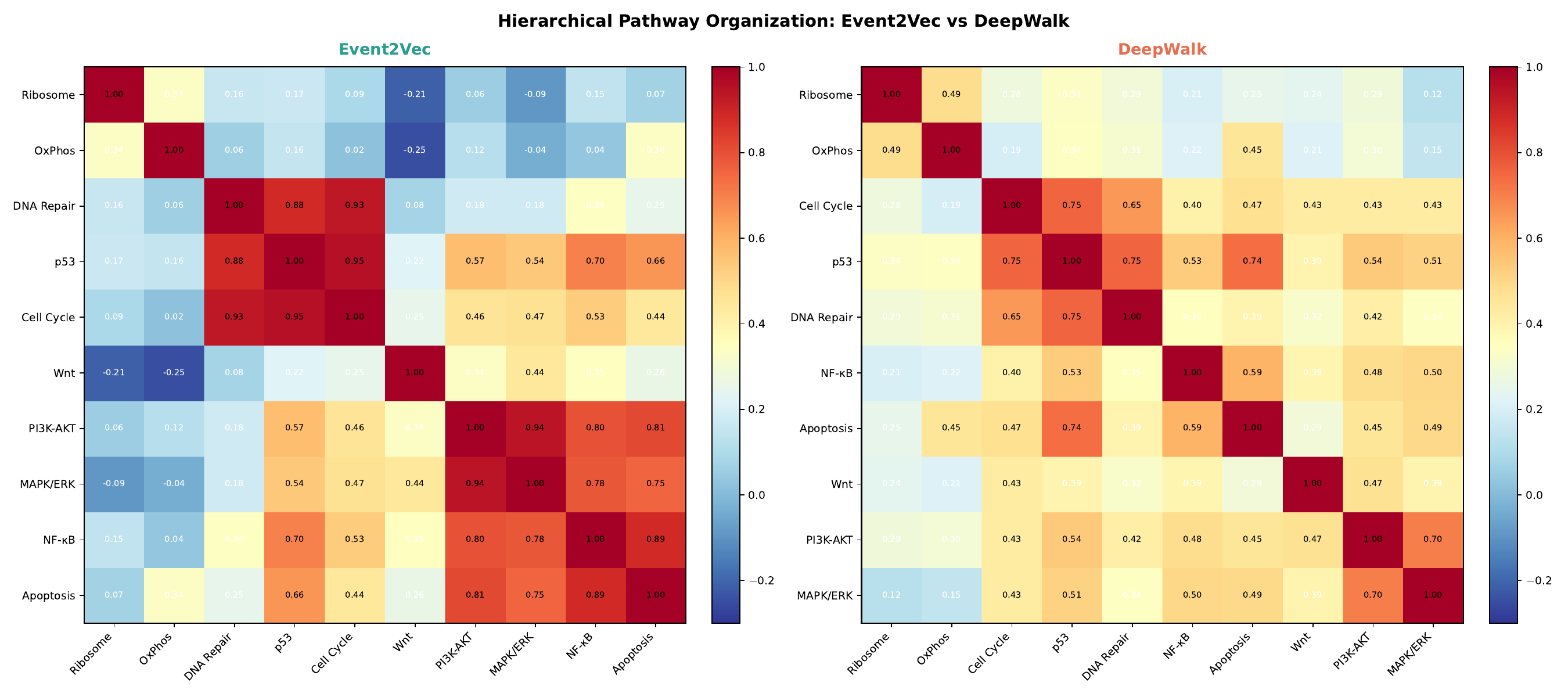}
\caption{\textbf{Hierarchical pathway organization.} Cosine similarity between pathway centroids, reordered by Ward clustering. Event2Vec (left) reveals three biologically coherent super-clusters: housekeeping (Ribosome, OxPhos), nuclear/genome (DNA Repair, p53, Cell Cycle), and signaling (PI3K-AKT, MAPK/ERK, NF-$\kappa$B, Apoptosis). Negative correlations (blue) sharply delineate housekeeping pathways from signaling cascades. DeepWalk (right) shows uniformly positive similarities without clear cluster boundaries, indicating weaker functional discrimination.}
\label{fig:heatmap}
\end{figure}

Clustering pathway centroids reveals clearer biological themes under Event2Vec (Figure~\ref{fig:heatmap}). Centroids are computed as the mean embedding of each pathway's member proteins, and cosine similarity matrices are reordered via Ward hierarchical clustering.

Event2Vec produces three distinct super-clusters with sharp boundaries. \textbf{Housekeeping} pathways (Ribosome, OxPhos) show high internal similarity (0.34) but low or negative similarity to all other pathways ($-0.25$ to $0.34$), reflecting their functional isolation from signaling processes. \textbf{Nuclear/genome} pathways (DNA Repair, p53, Cell Cycle) form a tight cluster (0.88--0.95 internal similarity), consistent with their shared role in maintaining genomic integrity. \textbf{Signaling} pathways (PI3K-AKT, MAPK/ERK, NF-$\kappa$B, Apoptosis) cluster together (0.75--0.94), reflecting extensive crosstalk between growth factor and stress response cascades.

DeepWalk's heatmap lacks this organization: all pairwise similarities are positive (0.12--0.75), cluster boundaries are diffuse, and the housekeeping pathways are not clearly separated from signaling. The absence of negative correlations indicates that DeepWalk places all pathways in a relatively uniform region of embedding space, without the functional discrimination that Event2Vec achieves.

\subsection{Embedding Drift Validates Additivity}

\begin{figure}[t]
\centering
\includegraphics[width=\linewidth]{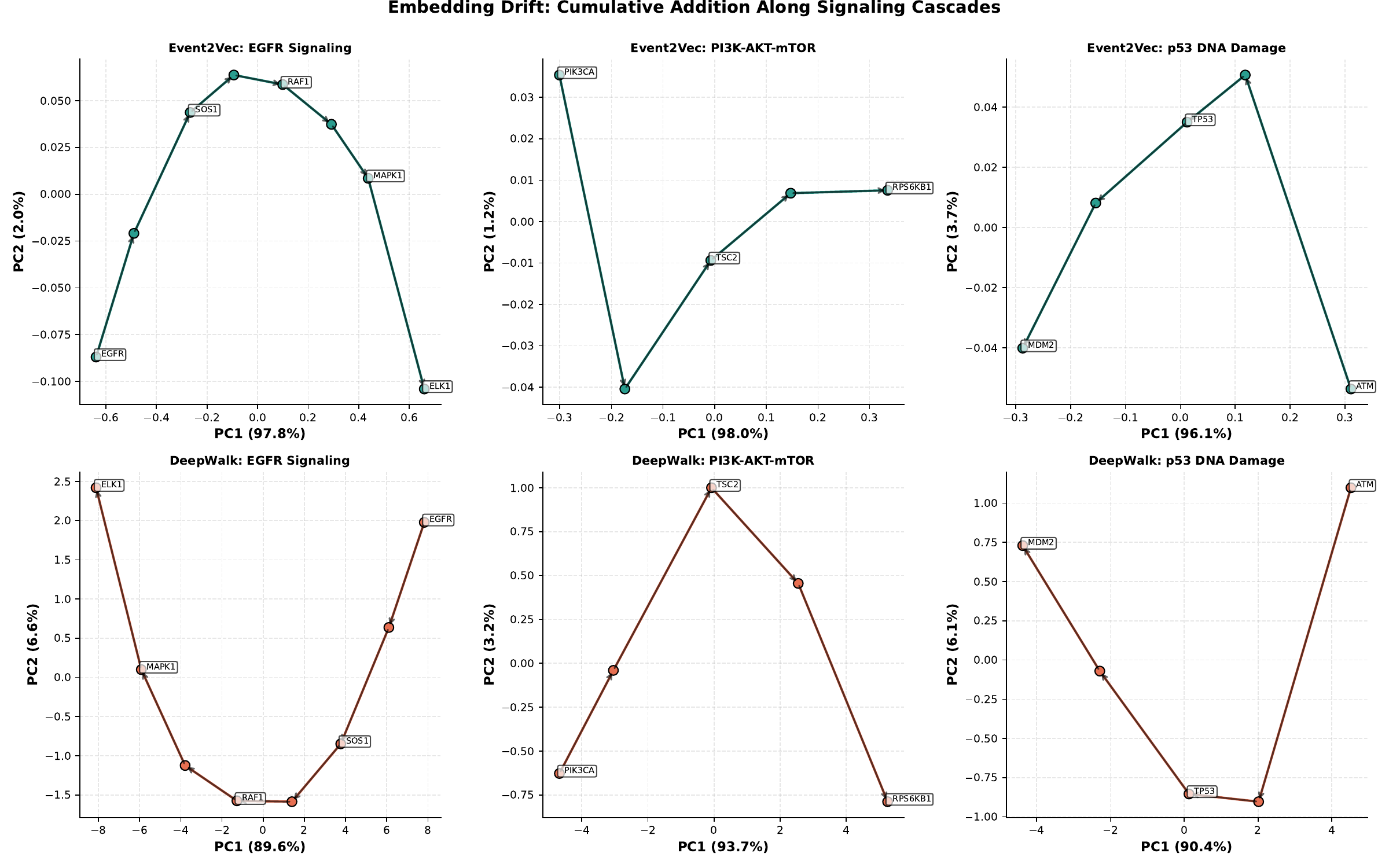}
\caption{\textbf{Embedding drift along signaling cascades.} Cumulative sums of protein embeddings ($h_t = \sum_{i=1}^{t} e_i$) are computed along three canonical pathways (EGFR signaling, PI3K-AKT-mTOR, p53 DNA damage) and projected via PCA. Event2Vec (top row) produces smooth, directional trajectories where PC1 captures 96--98\% of variance, indicating that sequential addition follows a coherent linear path. DeepWalk (bottom row) shows noisier trajectories with greater PC2 spread (PC1 explains only 89--94\% of variance), reflecting the absence of an additive constraint. This directly validates Event2Vec's compositional structure: the embedding of a signaling history approximates the sum of its constituent proteins.}
\label{fig:drift}
\end{figure}

Cumulative sums along signaling cascades follow smooth, directional trajectories under Event2Vec (PC1 explains 96--98\% of variance), while DeepWalk trajectories are noisier (PC1 explains 89--93\%) with more spread in PC2 (Figure~\ref{fig:drift}). This directly confirms that Event2Vec's additive constraint ($h_t = \sum_{i=1}^t e_i$) produces more coherent sequential representations.

\subsection{Summary Comparison}

Table~\ref{tab:comparison} summarizes results across all evaluation axes.

\begin{table}[t]
\centering
\small
\begin{tabular}{lcc}
\toprule
\textbf{Metric} & \textbf{Event2Vec} & \textbf{DeepWalk} \\
\midrule
\multicolumn{3}{l}{\textit{Compositional advantages (Event2Vec wins)}} \\
Pathway coherence (mean) & \textbf{0.870} & 0.648 \\
Coherence / random baseline & \textbf{30.2$\times$} & 2.9$\times$ \\
Relationship transfer (mean sim) & \textbf{0.966} & 0.650 \\
Top-1 analogy (EGFR:AKT1::INSR:?) & \textbf{KRAS} & AKT3 \\
Pathway cluster structure & \textbf{Sharp} & Diffuse \\
Embedding drift (PC1 variance) & \textbf{96--98\%} & 89--93\% \\
\midrule
\multicolumn{3}{l}{\textit{Shared / DeepWalk advantages}} \\
Norm--degree corr.\ ($r$) & $-0.627$ & $\mathbf{-0.801}$ \\
Hub centrality (fold closer) & \textbf{3.6$\times$} & 2.7$\times$ \\
Hub centrality ($p$-value) & $1.0\times10^{-9}$ & $\mathbf{1.6\times10^{-11}}$ \\
Drug target separation ($p$) & $8.4\times10^{-7}$ & $\mathbf{3.8\times10^{-8}}$ \\
\bottomrule
\end{tabular}
\caption{Comparison of Event2Vec and DeepWalk across all evaluation axes. Event2Vec shows large advantages in compositional tasks (arithmetic, transfer, clustering), while norm-based geometric properties are shared with or exceeded by the non-compositional baseline.}
\label{tab:comparison}
\end{table}

\section{Discussion}

The comparison with DeepWalk reveals a clear pattern: \textbf{compositional structure specifically benefits compositional tasks}. We discuss the advantages, shared properties, and limitations.

\textbf{Compositional advantages.} Event2Vec dramatically outperforms DeepWalk on tasks requiring relational reasoning: pathway coherence (30.2$\times$ vs 2.9$\times$ above random), relationship transfer (0.966 vs 0.650 similarity), and hierarchical pathway organization (sharp cluster boundaries vs diffuse structure). These gains directly validate the additive constraint: by enforcing $h_t = h_{t-1} + e_{s_t}$, Event2Vec learns functional \emph{transformations} as vectors, enabling the compositional arithmetic that DeepWalk cannot perform. The practical implication is that Event2Vec embeddings support biological reasoning (e.g., ``what plays the role of KRAS in insulin signaling?'') that skip-gram embeddings do not.

\textbf{Shared geometric properties.} The norm--degree anticorrelation ($r=-0.63$ for Event2Vec, $r=-0.80$ for DeepWalk) and drug target norm separation are not compositional signatures but shared skip-gram frequency effects. High-degree proteins appear more frequently in random walks and receive smaller norms under either model. We note that our earlier framing of norm as a ``hyperbolic interpretation'' specific to Event2Vec was overstated; this property is better understood as an artifact of the skip-gram training shared by both methods. That said, Event2Vec does achieve sharper hub--periphery separation (3.6$\times$ vs 2.7$\times$ fold-change), suggesting the additive constraint provides modest additional centrality structure.

\textbf{Limitations.} Event2Vec cannot capture context-dependent or multiplicative interactions. Cross-compartment analogies fail (the model treats all interactions in a single space). The evaluation, while expanded beyond initial examples, still covers a limited set of pathways and relationships. Future work should validate against comprehensive databases (PhosphoSitePlus, KEGG) using systematic hits@k metrics.

\textbf{Applications.} The compositional advantages suggest specific use cases: function annotation via arithmetic, pathway crosstalk identification via heatmap similarities, and drug repurposing via relationship transfer. For tasks depending primarily on network centrality (e.g., hub identification, drug target ranking by degree), simpler methods may suffice.

\section{AI-Assisted Research Methodology}

\subsection{Overview of Collaborative Approach}

This work represents a collaborative effort between human researchers and Claude Sonnet 4.5 (Anthropic), where AI served as an interpretive assistant for analyzing Event2Vec embeddings of protein networks. The collaboration leveraged complementary strengths: human expertise in study design, method implementation, and statistical validation, combined with AI's capacity for rapid pattern recognition and biological contextualization. All computational results are fully reproducible from provided code, while biological interpretations represent AI-assisted hypotheses that benefited from the model's broad knowledge of systems biology.

\subsection{Division of Responsibilities}

The research workflow divided naturally into computational and interpretive components. Human researchers designed the study, recognizing that Event2Vec's compositional structure could capture functional relationships in sequential protein interaction data. We implemented Event2Vec training on the STRING interactome, developed six evaluation strategies based on our understanding of additive embeddings, and wrote all analysis code. We generated quantitative results for pathway coherence, network centrality, and geometric properties, then created visualizations to explore these patterns.

Claude's role began after computational analysis was complete. We provided the AI with our quantitative findings, including pathway similarity matrices, embedding distributions, and statistical summaries. Claude proposed specific biological examples to test our hypotheses: selecting eight hub proteins for centrality analysis, identifying patterns in pathway clustering that suggested three functional themes (signaling, nuclear/genome, and housekeeping), and recommending four protein arithmetic tests to demonstrate compositional learning. The AI also assisted substantially with manuscript preparation, transforming technical results into accessible scientific narrative and improving clarity throughout. Claude also recommended the DeepWalk baseline comparison, which proved essential for distinguishing compositional effects from general skip-gram properties.
Table~\ref{tab:ai-contributions} summarizes this division of labor.

\begin{table}[h]
\centering
\small
\begin{tabular}{lcc}
\toprule
\textbf{Research Activity} & \textbf{Human} & \textbf{AI} \\
\midrule
Study design \& implementation & Complete & None \\
Evaluation strategy development & Complete & None \\
Statistical analysis \& validation & Complete & None \\
Baseline comparison design & Joint & Joint \\
Example \& test case selection & Verification & Complete \\
Biological pattern identification & Verification & Proposal \\
Manuscript preparation & Editing & Editing \\
\bottomrule
\end{tabular}
\caption{Division of labor between human researchers and AI assistant.}
\label{tab:ai-contributions}
\end{table}

\subsection{Workflow and Validation}

The collaboration proceeded in a focused session. After completing all computational analyses independently, we presented Claude with our results alongside relevant biological context (protein annotations, pathway definitions, known functional relationships). Claude examined pathway clustering patterns and proposed the three-theme organization, suggested specific hub proteins whose centrality we could test, and recommended protein arithmetic examples that would demonstrate compositional structure. For each AI suggestion, we computed the corresponding statistics and verified quantitative accuracy.

This approach proved efficient: Claude's biological knowledge helped identify the most informative examples from our high-dimensional embedding space, while our statistical verification ensured all claims rested on solid computational ground. Crucially, the AI's recommendation to add a DeepWalk baseline led to a more nuanced and honest characterization of Event2Vec's strengths, revealing that some geometric properties we initially attributed to compositionality are in fact shared skip-gram effects.

\subsection{Strengths and Limitations of the Approach}

The collaboration successfully combined computational rigor with biological interpretation. All statistical findings are independently verifiable and reproducible from our code. The DeepWalk baseline comparison strengthened the paper by identifying which properties are genuinely compositional and which are shared across skip-gram methods.

The biological interpretations---while plausible and statistically supported---represent AI-generated hypotheses rather than independently validated findings. We did not cross-reference Claude's three-theme classification against established pathway hierarchies (KEGG, Reactome, GO), nor did we systematically evaluate the full distribution of protein arithmetic similarities beyond the reported tests. Future work could strengthen these interpretations through expert consultation or systematic database cross-referencing.

\subsection{Contribution and Positioning}

This work makes both a technical and methodological contribution. Technically, we show that enforcing compositional structure in protein network embeddings yields specific advantages for relational reasoning (arithmetic, transfer, pathway clustering) while some geometric properties (norm--degree correlation, drug target separation) are shared with non-compositional baselines. Methodologically, we demonstrate how AI can accelerate the interpretation of complex biological embeddings, while highlighting the importance of maintaining clear boundaries between computational findings and biological hypotheses.

We position the biological interpretations as AI-assisted hypotheses worthy of further investigation. By providing our embeddings and code openly, we invite the community to test these hypotheses and explore alternative interpretations.

\subsection{Overview of AI Involvement}

This work utilized Claude Sonnet 4.5 (Anthropic) as an interpretive assistant for analyzing Event2Vec embeddings of protein networks. The AI's role was \textbf{not as co-discoverer but as interpreter}: humans designed the study, implemented all methods, and generated all quantitative results. Claude assisted in (1) selecting specific biological examples to test, (2) interpreting statistical patterns in biological terms, (3) recommending the baseline comparison, and (4) manuscript preparation. All core scientific claims are grounded in reproducible computational results, while biological interpretations represent AI-assisted hypotheses requiring independent validation.

All embeddings, code, and quantitative results will be openly provided to enable such independent evaluation upon publication. 

\section{Conclusion}

Event2Vec learns biologically meaningful protein embeddings through enforced compositional structure. Compared to a DeepWalk baseline on identical data, Event2Vec shows large advantages in pathway coherence (30.2$\times$ vs 2.9$\times$ above random), functional arithmetic (0.966 vs 0.650 mean transfer similarity), and hierarchical pathway organization. However, norm-based geometric properties such as the degree anticorrelation are shared with the non-compositional baseline. The key insight is that \textbf{compositional constraints specifically benefit compositional tasks}: relationship transfer, functional analogy, and pathway clustering all require the additive structure that Event2Vec enforces but DeepWalk lacks.

\bibliographystyle{iclr2026_conference}
\bibliography{references}

\newpage
\appendix
\section{Appendix: Extended Analyses}

\subsection{Geodesic Path Reconstruction}

Linear interpolation between protein pairs reconstructs plausible signaling cascades. For EGFR$\rightarrow$TP53, intermediate points identify: ERBB2, STAT3, NFKB1, RELA, CCND1, TP53, USP7 (mean similarity 0.97). For TNF$\rightarrow$CASP3: TLR4, IL1B, SOCS3, ANXA5, CASP3, CASP9, BCL2L11 (0.98). For BRCA1$\rightarrow$RAD51: MDC1, BRCA2, CHEK2, RAD51, ORC1, ERCC4, FANCI (0.99).

These paths suggest mechanistic intermediates for experimental validation. The high similarities ($>$0.96) indicate smooth geometric transitions matching biological sequences.

\subsection{Compositional Algebra Tests}

\textbf{Relationship transfer}: Functional transformations (Receptor$\rightarrow$Effector, TF$\rightarrow$Target, Kinase$\rightarrow$Substrate) transfer across protein families with 0.94--0.98 similarity under Event2Vec, compared to 0.58--0.74 under DeepWalk. This is the most informative test of compositional structure, as it requires learned transformations to generalize to new proteins.

\textbf{Pathway decomposition}: All 10 pathway centroids reconstruct perfectly from member proteins (centroid = $\sum$ proteins $/n$, similarity$>$0.999), validating linear combinability. Note: this property holds by definition for any embedding and does not distinguish between methods.

\subsection{Per-Pathway Coherence Comparison}

\begin{table}[h]
\centering
\small
\begin{tabular}{lccc}
\toprule
\textbf{Pathway} & \textbf{Event2Vec} & \textbf{DeepWalk} & \textbf{$\Delta$} \\
\midrule
Ribosome & 0.993 & 0.784 & +0.209 \\
DNA Repair & 0.966 & 0.683 & +0.283 \\
MAPK/ERK & 0.952 & 0.641 & +0.311 \\
NF-$\kappa$B & 0.950 & 0.638 & +0.312 \\
OxPhos & 0.938 & 0.718 & +0.220 \\
Cell Cycle & 0.886 & 0.687 & +0.199 \\
Wnt & 0.844 & 0.679 & +0.166 \\
Apoptosis & 0.810 & 0.648 & +0.162 \\
PI3K-AKT & 0.685 & 0.502 & +0.183 \\
p53 & 0.681 & 0.505 & +0.176 \\
\midrule
Mean & \textbf{0.870} & 0.648 & +0.222 \\
Random baseline & 0.029 & 0.223 & --- \\
\bottomrule
\end{tabular}
\caption{Per-pathway coherence for Event2Vec vs DeepWalk. Event2Vec exceeds DeepWalk on all 10 pathways.}
\label{tab:per-pathway}
\end{table}

\subsection{Systematic Relationship Transfer Results}

Across 15 transfer tests (5 query proteins $\times$ 3 relationship types), Event2Vec predictions are both higher-similarity and more biologically specific. DeepWalk's Receptor$\rightarrow$Effector predictions collapse to AKT3 for 4/5 queries, indicating it returns the nearest neighbor to the query rather than completing the relational transformation. Full results are provided in the supplementary code.

Random walks: 10 per protein, length 15, uniform sampling. Event2Vec training: Adam optimizer, learning rate 0.001, batch size 32, 100 epochs, reconstruction weight $\lambda=0.5$. Total training time: $\sim$30 minutes on single T4 GPU. DeepWalk baseline: gensim Word2Vec, skip-gram, identical dimensionality (64), window (5), and epochs (100).

Pathway definitions from KEGG and Reactome databases. Hub proteins selected as top-8 by degree centrality. Drug targets: FDA-approved small molecule targets with $\geq$50 connections in STRING.

\end{document}